# Some Improvements on Deep Convolutional Neural Network Based Image Classification

**Andrew G. Howard**
Andrew Howard Consulting
Ventura, CA 93003
*andrewgeraldhoward@gmail.com*

**Abstract**

We investigate multiple techniques to improve upon the current state of the art deep convolutional neural network based image classification pipeline. The techniques include adding more image transformations to the training data, adding more transformations to generate additional predictions at test time and using complementary models applied to higher resolution images. This paper summarizes our entry in the Imagenet Large Scale Visual Recognition Challenge 2013. Our system achieved a top 5 classification error rate of 13.55% using no external data which is over a 20% relative improvement on the previous year's winner.

## 1    Introduction

Deep convolutional neural networks have recently been substantially improving upon the state of the art in image classification and other recognition tasks [2,6,10]. Since their introduction in the early 1990s [7], convolutional neural networks have consistently been competitive with other techniques for image classification and recognition. Recently, they have pulled away from competing methods due the availability of larger data sets, better models and training algorithms and the availability of GPU computing to enable investigation of larger and deeper models.

The Imagenet Large Scale Visual Recognition Challenge (ILSVRC) [8] is a venue for evaluating what the current state of the art for image classification and recognition is. It is large dataset of 1.2 million images with 1000 classes that are a subset of the Imagenet dataset [3]. The 2012 competition demonstrated a large step forward for convolutional neural networks where a convnet based system [6] soundly beat the competing methodology based on Fisher Vectors [4] by roughly 10% in absolute terms. The convolution neural network based system was made up of an ensemble of deep, eight layer networks. It also incorporated important features such as pooling and normalizing layers, data transformations to expand the training data size, data transformations at test time, and drop out [5] to avoid over fitting.

We investigate useful additions to the winning system from 2012 [6] and improve upon its results by 20% in relative terms. This paper summarizes our entry in the 2013 ILSVRC which achieved a 13.55% top 5 error rate compared to the previous year's 16.4% top 5 error rate. The winning entry, Clarifai, which is partially detailed in [10], achieved a top 5 error rate of 11.74%. The methods outlined in this paper should be able to improve upon it and other convolutional neural network based systems.

Our models are based on the 2012 winning system [6] and use the code provided at http://code.google.com/p/cuda-convnet as a starting point. Our model structure is identical with the exception that the fully connected layers are twice as big. It turns out, that this change does not improve the top 5 error rate. We use the same training methodology of training the net until the validation error plateaus and reducing the step size by 10 at each plateau. Additional changes are detailed in the follow sections.

The paper is organized as follows. Section 2 describes the additional transformations used to increase the effective number of training examples. Section 3 describes the additional transformations used at test time to improve prediction and a method to reduce the total number of predictions to a manageable size. Section 4 describes complementary high resolution models. Section 5 reports the results of our system and we conclude with a summary and discussion.

## 2    Additional Data Transformations for Training

Deep neural networks are greatly improved with the addition of more training data. When more training data is not available, transformations to the existing training data which reflect the variation found in images can synthetically increase the training set size. In the previous Imagenet classification system [6], three types of image transformations were used to augment the training set. The first was to take a randomly located crop of 224x224 pixels from a 256x256 pixel image capturing some translation invariance. The second was to flip the image horizontally to capture the reflection invariance. The final data transformation was to add randomly generated lighting which tries to capture invariance to the change in lighting and minor color variation. We add additional transformations that extend the translation invariance and color invariance.

### 2.1    Extending Image Crops into Extra Pixels

Previously [6], translated image crops of 224x224 pixels were selected from a training image of 256x256. The 256x256 image was generated by rescaling the largest image dimension to 256 and then cropping the other side to be 256. This results in a loss of information by not considering roughly 30% of the pixels. While the cropped pixels are most likely less informative than the middle pixels we found that making use of these additional pixels improved the model.

To use the whole image, we first scale the smallest side to 256 leaving us with a 256xN or Nx256 sized image. We then select a random crop of 224x224 as a training image. This yields a large number of additional training examples and helps the net learn more extensive translation invariance. Figure 1 shows a comparison of a square cropped image versus using the full image. The square training image of the cat will never generate training examples with a tail in it or with both ears compared to selecting crops from the full image.

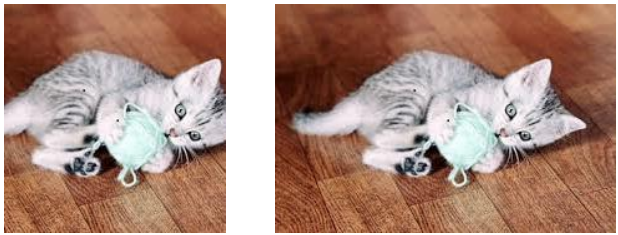

Figure 1: Even well centered images when cropped lose information like the cat's ear and tail compared to the full image on the right. We select training patches from the full image to avoid loss of information.

### 2.2    Additional Color Manipulations

In addition to the random lighting noise that has been used in previous pipelines [6], we also add additional color manipulations. We randomly manipulate the contrast, brightness and color using the python image library (PIL). This helps generate training examples that cover the span of image variations helping the neural network to learn invariance to changes in these properties. We randomly choose an order for the three manipulations and then choose a number between 0.5 and 1.5 for the amount of enhancement (a setting of 1 leaves the image unchanged). After manipulating the contrast, brightness and color, we then add the random lighting noise similar to [6].

## 3    Additional Data Transformations for Testing

Previous methods combined the predictions of 10 image transformations into a final prediction. They used the central crop and the four corners as well as the horizontal flip of these five. We found that predicting at three different scales improved the joint prediction. We also made predictions on three different views of the data capturing the extra pixels that are previously cropped out. The combination of 5 translations, 2 flips, 3 scales, and 3 views leads to 90 predictions which slow predictions down by almost an order of magnitude. To rectify this, we show that a simple greedy algorithm can choose a subset of 10 transforms that predicts almost as well as all 90 and a subset of 15 that predicts slightly better.

### 3.1    Predictions at Multiple Scales

Images contain useful predictive elements at different scales. To capture this we make predictions at three different scales. We use the original 256 scale as well as 228 and 284. Note that when scaling an image up, it is important to use a good interpolation method like bicubic scaling and not to use anti aliasing filters designed for scaling images down. When scaling down, anti aliasing seems to help a little bit but in practice we used bicubic scaling for up scaling and down scaling.

### 3.2    Predictions with Multiple Views

In order to make use of all of the pixels in the image when making predictions, we generate three different square image views. For an 256xN (Nx256) image, we generate a left (upper), center, and right (lower) view of 256x256 pixels and then apply all crops, flips and scales to each of these views. Figure 2 demonstrates how the three views are constructed.

Table 1 shows the effect of using the new training and testing transforms compared to the previous baseline result. It also shows that the new model architecture with doubled fully connected layers does not improve the top 5 error rate.

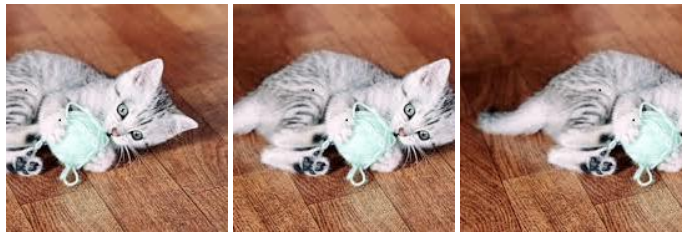

Figure 2: We generate predictions based on three different square views of the image to incorporate all of the pixels and to take into account differing image sizes.

|  | Val Top-1 | Val Top-5 |
|---|---|---|
| Krizhevsky et al Single Convnet [6] | 40.7% | 18.2% |
| New Training, Test Transforms | 37.5% | 15.9% |
| Double FC + New Training, Test Transforms | 37.0% | 15.8% |

Table 1: Results for new training and testing transforms using the architecture of Krizhevsky et al and the same model with double the size for all fully connected layers. The larger fully connected layers do not improve the top 5 validation error substantively.

### 3.3    Reducing the Number of Predictions

Our final combination of 5 crops, 2 flips, 3 scales and 3 views yields a combination of 90 predictions per model. This is almost an order of magnitude larger than the 10 predictions that were used previously and may cause an unacceptable delay in real world applications. We used a simple greedy algorithm to reduce the number of predictions to an acceptable number.

The simple greedy algorithm starts with the best prediction and at each step adds another prediction until there is no additional improvement. This algorithm finds that the first 10 predictions are almost as accurate as using all 90 which would have the same run time as previous methods that also only use 10 predictions. It is also able to find a slightly improved combination using 15 predictions that improves on the 90 prediction baseline. Figure 2 shows a plot of accuracy as more predictions are added.

This simple greedy algorithm is easy to implement, quick to run and has no parameters to tune. It avoids some of the pitfalls inherent in various weight based learning methods for combining predictions. Because most of the predictions are very similar (almost collinear), methods such as stacking [9] or similar algorithms tend to have trouble and can create large weights of opposing sign. Penalization based method (l1, l2 etc) can help but not completely mitigate this effect. Table 2 shows the effect of the new test transformations and the results of the greedy algorithm. Figure 3 shows the progression of the greedy algorithm as it adds models until convergence.

|  | Val Top-1 | Val Top-5 |
|---|---|---|
| 10 Predictions: 5 Crops, 2 Flips | 39.1% | 17.4% |
| 30 Predictions: 5 Crops, 2 Flips, 3 Scales | 38.3% | 16.7% |
| 30 Predictions: 5 Crops, 2 Flips, 3 Views | 37.7% | 16.4% |
| 90 Predictions: 5 Crops, 2 Flips, 3 Scales, 3 Views | 37.1% | 15.9% |
| 10 Predictions: Greedy | 37.2% | 16.0% |
| 15 Predictions: Greedy | 37.1% | 15.9% |

Table 2: This table shows results for additional test time transformations and combinations with a greedy algorithm.

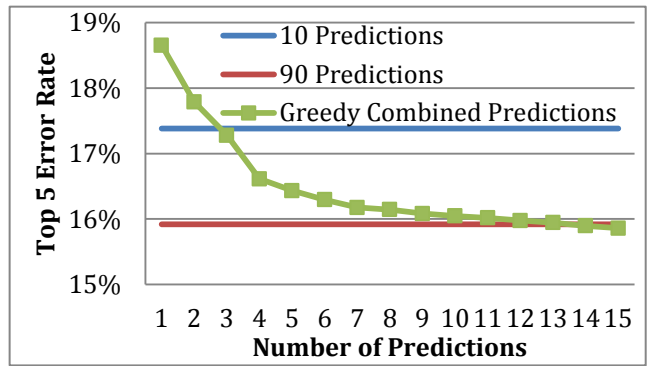

Figure 3: This figure shows the accuracy of the greedy selection algorithm as it adds more predictions compared to the baseline 10 predictions and the full 90 predictions.

## 4 Higher Resolution Models

Objects in images can exist at different scales. In the previous section we run our trained network on scaled up and scaled down versions of the image. However, once the image is scaled up too much, the trained network no longer performs well. To search for objects in images in scaled up higher resolution images we need to retrain the network at that scale. In practice, previously trained models can be used to initialize higher resolution models and cut training time down substantially to 30 epochs from 90 epochs. Higher resolution models are complementary to base models and a single high resolution model is as valuable as four additional base models as shown in Table 3. Complementary models have been found to be valuable in other large scale data contests such as the Netflix Prize [1].

### 4.1 Model Details

In previous sections, models are trained on 224x224 patches cropped from 256xN (Nx256) images. For the higher resolution models, we wish to train on larger images using the same model structure. We build models on 448xN (Nx448) images using crops of 224x224. This ensures that at test time, the four corner crops do not overlap. In practice it may not be practical to store images at multiple resolutions or they may not be available. So to reuse 256xN (Nx256) training images we take 128x128 sized patches and scale them up to be 224x224. This simulates using 224x224 patches in a 448xN (Nx448) image without needing access to separate training images.

We use an identical model structure and training methods as the lower resolution models. Because the model is identical in structure, we can initialize the higher resolution models with fully trained low resolution models. This allows us to train the new model in 30 epochs rather than the previous 90 epochs. We start with a step size of 0.001 and divide the step size by 10 once the validation score plateaus. We reduce the step size twice before convergence. Because there are effectively more training patches due to the smaller patch size relative to image size, drop out is not as important. In practice we use drop out for the initial training and the first step size reduction to 0.0001, we then turn off drop out to finish the training at 0.0001. Then the model is finished training with drop out turned off and the step size reduced to 0.00001. This gives better results than training with drop out alone or without drop out.

When making predictions on new images, we can make use of more image patches. Previously 5 crops (or translations) are used when making predictions. Because these crops have less overlap for the higher resolution model, we find it useful to increase this to 9 crops by adding the middle top, middle bottom, left middle and right middle. This brings the total number of predictions per model up to 162 (9 crops, 2 flips, 3 scales, 3 views) making the greedy selection algorithm from section 3.3 very important.

|  | Val Top-1 | Val Top-5 |
|---|---|---|
| **One Base Net** | 37.0% | 15.8% |
| **Two Base Nets** | 35.9% | 15.1% |
| **One High Resolution Net** | 36.8% | 16.2% |
| **One Base Net + One High Resolution Net** | 34.9% | 14.5% |
| **Five Base Nets** | 35.2% | 14.5% |

Table 3: The combination of base model and high resolution model is better than five base models for top-1 classification and equivalent for top-5 classification.

## 5 Results

The final image classification system submitted to ILSVRC2013 was composed of 10 neural networks made up of 5 base models and 5 high resolution models and had a test set top 5 error rate of 13.6%. This is an improvement on the previous state of the art of 16.4% but short of the best result of 11.7%. The methods described in this paper should be able improve on the current state of the art of 11.7%. Results are summarized in Table 4.

|  | Val Top-1 | Val Top-5 | Test Top-5 |
|---|---|---|---|
| **Krizhevsky et al [] Five Nets** | 38.1% | 16.4% | 16.4% |
| **Five Base Nets** | 35.2% | 14.5% | - |
| **Five High Resolution Nets** | 35.3% | 15.1% | - |
| **Five Base + Five High Resolution Nets** | 33.7% | 13.7% | 13.6% |
| **Clarifai** | - | - | 11.7% |

Table 4: This table shows a comparison of the proposed methods to the previous year's winner Krizhevsky et al [6] and current year's winner Clarifai. The improvements proposed could also improve on the Clarifai system.

# 6    Conclusion

In this paper we showed a number of ways to improve neural network based image classification systems. We first showed some new useful image transformations to increase the effective size of the training set. These were based on using more of the image to select training crops and additional color manipulations. We also showed useful image transformations for generating testing predictions. We made predictions at different scales and generated predictions on different views of the image. These additional predictions can slow down the system so we showed a simple greedy algorithm that reduces the number of predictions needed. Finally, we showed an efficient way to train higher resolution models that generate useful complementary predictions. A single base model and a single high resolution model are as good as 5 base models. These improvements to the image classification pipeline are easy to implement and should be able to improve other convolutional neural network based image classification systems.